\documentclass[letterpaper, 10 pt, journal, twoside]{IEEEtran}

\usepackage{booktabs}
\usepackage{hyperref}
\usepackage{booktabs}
\usepackage{graphicx} 
\usepackage{epsfig} 
\usepackage{amsmath} 
\usepackage{amssymb}  
\usepackage{gensymb}
\usepackage{xcolor}
\usepackage{array}
\usepackage{rotating}
\usepackage{tabularray}
\definecolor{blue}{rgb}{0,0,0}
\usepackage{pdfpages}
\usepackage{dblfloatfix}
\usepackage{multirow}
\usepackage[switch]{lineno}
\usepackage{cite}

\begin{document}

\title{A Novel Transformer-Based Method for Full Lower-Limb Joint Angles and Moments Prediction in Gait Using sEMG and IMU data}

\author{F.H. Daryakenari$^{1}$, T. Farizeh$^{*2}$
\thanks{$^1$Faculty of Engineering and IT, the University of Melbourne, Victoria, Australia. {\tt\footnotesize f.haghgoodaryakenari@unimelb.edu.au}; $^2$Faculty of Caspian College of Engineering, University of Tehran, Tehran, Iran. {\tt\footnotesize tara.farizeh@ut.ac.ir}}
\thanks{Digital Object Identifier (DOI): see top of this page.}
}


\maketitle

\begin{abstract}
This study presents a transformer-based deep learning framework for the long-horizon prediction of full lower-limb joint angles and joint moments using surface electromyography (sEMG) and inertial measurement unit (IMU) signals. Two separate Transformer Neural Networks (TNNs) were designed: one for kinematic prediction and one for kinetic prediction. The model was developed with real-time application in mind, using only wearable sensors suitable for outside-laboratory use. Two prediction horizons were considered to evaluate short- and long-term performance. The network achieved high accuracy in both tasks, with Spearman correlation coefficients exceeding \(\rho\) = 0.96 and \(R^2\) scores above 0.92 across all joints. Notably, the model consistently outperformed a recent benchmark method in joint angle prediction, reducing RMSE errors by an order of magnitude. The results confirmed the complementary role of sEMG and IMU signals in capturing both kinematic and kinetic information. This work demonstrates the potential of transformer-based models for real-time, full-limb biomechanical prediction in wearable and robotic applications, with future directions including input minimization and modality-specific weighting strategies to enhance model efficiency and accuracy.

\end{abstract}

\IEEEpeerreviewmaketitle
\section{Introduction}
\IEEEPARstart{A}{crucial} requirement in developing real-world systems—especially those that involve repetitive tasks—is optimization. Without an optimized system, we risk excessive energy consumption, increased physical or computational effort, and ultimately higher operational costs, all of which are undesirable. However, achieving such optimization requires a foundational step: analyzing the system’s dynamics throughout task execution. When the system is intended for human use—such as in rehabilitation, athletic performance enhancement, or elderly care—this analysis becomes even more critical due to strict safety requirements and the need for individualized adaptation. Despite the fact that humans perform a wide range of daily motions unconsciously, analyzing these movements remains challenging. This is largely due to the biomechanical complexity of the human body and the difficulty of estimating external forces and motions, even in fundamental activities such as gait. These challenges underscore the need for a reliable and real-time method to extract both kinematic and kinetic information during movement.

The significance of human motion data and the methods for its accurate extraction have been extensively studied across a wide range of disciplines \cite{Buchanan2005, Crenna2011, Chowdhury2013}. Early research primarily tried to explain the underlying mechanism of human locomotion such as the role of joint forces and moments in muscle-tendon dynamics. This foundational knowledge has been a key asset in rehabilitation, orthopedic assessment, and enhancing sports performance \cite{Hov1990, Maganaris2004, Chowdhury2013, Ceseracciu2014, Kanko2021}. Later, human motion analysis gained prominence in the entertainment industry—including film, animation, and video games—by enabling the creation of realistic and expressive character movements \cite{Thalmann1997, Herda2000, Liu2006, Aristidou2008, Aristidou2013}. More recently, its relevance has expanded into robotics, where detailed information about joint kinematics and kinetics has become critical for the control of prosthetic limbs, exoskeletons, and assistive devices \cite{Buchanan2004, Aristidou2008, Aristidou2013, Gabbasov2015, Wang2021}.

Early motion analysis methods relied heavily on mathematical and statistical models, requiring significant computational efforts, which made them unsuitable for real-time applications. Moreover, due to the simplifications inherent in these models, they failed to accurately mimic human movements. 
Subsequently, methods based on infrared (IR) optical motion capture systems were developed and refined \cite{Herda2000, Liu2006, Aristidou2008, Aristidou2013}. Achieving accurate data acquisition with these systems requires multiple IR cameras, which makes this approach costly and impractical for widespread use. Additionally, the recorded data from IR cameras are processed using inverse kinematics and kinetics solvers to calculate motion parameters. Since this data transformation process is not instantaneous, these systems are unsuitable for real-time applications. Their dependence on a laboratory setup further restricts their use to controlled environments. Despite these drawbacks, the high accuracy of marker-based motion capture systems makes them a reliable source for ground-truth data, particularly for validating other motion analysis methods.

To address the limitations of optical motion capture systems and enable real-world human motion analysis, sensor-based methods have been developed \cite{Willemsen1990, MoeNilssen2004, Selles2005, Picerno2017}. Early attempts, such as those by \cite{Goulermas2005} and \cite{Findlow2008}, aimed to create a natural setting for measuring human movement and performance. They employed a regression neural network (NN) to predict hip, knee, and ankle kinematics in the sagittal plane using angular velocity and linear acceleration data from motion sensors attached to the foot and shank. Although conventional NN regressors were groundbreaking, they have since been replaced by more advanced models with greater capability to predict sequential data. To validate their predictions, these early works relied on optical motion capture systems as the gold standard.

Today, Inertial Measurement Units (IMUs) have emerged as a promising alternative to record motion data \cite{Lim2020, Weygers2020, Niswander2020, Pitt2020, Majumder2021, McGrath2022, Carcreff2022}. Their portability and ability to record data outside laboratory settings make them a practical substitute for optical motion capture systems. IMU data have proven reliable for extracting joint kinematics \cite{Versteyhe2020, Niswander2020, Pitt2020, LoraMillan2021, Sy2021, Majumder2021, McGrath2022, Potter2022, Carcreff2022}, joint kinetics \cite{Lee2020, Lee2022, Krishnakumar2024}, or both \cite{Dorschky2019, Mundt2020a, Mundt2020b, Stanev2021, Yi2021, Moghadam2023, Dimitrov2023}. This versatility and practicality position IMUs as a transformative tool in motion analysis. IMU data shows promising results in joint kinetic estimation and prediction \cite{Dorschky2019, Mundt2020a, Mundt2020b, Stanev2021, Yi2021, Moghadam2023, Dimitrov2023}. But, its inherently kinematic nature limits its standalone performance in accurately estimating joint kinetics. To improve accuracy in joint kinetics estimation, IMU data can be fused with other sensory data that inherently contain kinetic information. This fusion could enhance the prediction of kinetic data, especially in addressing non-periodic or perturbation-induced movements.

Surface electromyography (sEMG) data have long been studied as a complementary source for joint kinetics and muscle force estimation \cite{Ardestani2014, Brantley2017, Chen2018, Bi2019, Xiong2019, Huang2019, Liang2021, Sitole2023, Yaakoubi2023}. It provides better performance in kinetic analysis compared to kinematics \cite{Olney1985, Koo2005, Shao2009, Camargo2022}, but has also been explored for joint kinematics estimation or prediction, albeit with relatively lower accuracy \cite{Chen2018, Bi2019, Huang2019, Xiong2019, Liang2021, Rabe2022, Brantley2017, Koike1995, Lloyd2003, Zhang2013, Sitole2023}. Fusing IMU and sEMG data offers the potential to create a comprehensive dataset for joint kinematics and kinetics estimation or prediction. IMU data serves as the primary source for kinematics and a supplementary source for kinetics, while sEMG provides the inverse. One notable advantage of sEMG is its preactivation characteristic \cite{Merletti2020}, which reveals muscle activation before the actual movement. This feature is particularly advantageous in predictive applications, such as assistive robotics or tasks requiring rapid, precise actions. Based on this complementary relationship, we hypothesize that combining IMU and sEMG signals provides sufficient information for accurate joint angle and moment prediction. To realize this, it is essential to develop a mapping between fused sensory data and joint kinematics and kinetics, similar to the approach pioneered by \cite{Goulermas2005} and \cite{Findlow2008} using neural networks.

Several approaches have been proposed to map inertial data to joint kinematics or kinetics, but many struggle with real-time performance and generalizability. Kalman Filter (KF)-based methods, including Extended Kalman Filters (EKFs), are computationally efficient \cite{Baghdadi2018, Abdelhady2019, Teufl2019, Sy2020, LoraMillan2021, Sy2021, Potter2022}, but their reliance on linear assumptions, Gaussian noise, and short temporal windows makes them ill-suited for modeling the nonlinear and dynamic nature of human joint motion. They also require precise system and noise modeling, which is often difficult and error-prone. Similarly, musculoskeletal models (MSMs) offer high biomechanical fidelity by incorporating anatomical parameters \cite{Dorschky2019, McGrath2022}, but their subject-specific nature and computational demands limit their scalability and real-time applicability. The complexity of tuning and solving these models remains a major barrier for use in interactive scenarios such as human-robot interaction.

Recent advances in Artificial Intelligence (AI), particularly Deep Learning (DL), have enabled the modeling of complex, nonlinear systems such as human motion. Neural Networks (NNs), especially sequential architectures like Long Short-Term Memory (LSTM) and encoder–decoder models, are well-suited for time-series prediction due to their ability to capture temporal dependencies. Several studies have applied DL techniques to predict joint kinematics and kinetics from IMU \cite{Marimon2024, Hossain2023, Renani2021, Hernandez2021, Chen2021} or sEMG data \cite{Kaya2024, Zhang2023, Song2023, Shi2023, Truong2023, Foroutannia2022, Wu2021, Zhang2021, Wang2025}. Although most of these methods rely on a single modality, limiting generalization. Notably, Wang et al. \cite{Wang2021} introduced a fused sEMG–IMU model using a Temporal Convolution Network–Bidirectional LSTM for joint angle prediction but excluded joint moment estimation. To the best of the authors’ knowledge, the only recent attempt to estimate both kinematics and kinetics from fused signals is by Mohammadi Moghadam et al. \cite{Moghadam2023}, who showed that Convolutional Neural Networks (CNNs) outperformed traditional ML methods. However, their reliance on offline preprocessing and CNNs’ limited temporal modeling made their approach unsuitable for real-time deployment. These limitations highlight the need for a unified, multimodal method capable of capturing both spatial and temporal features in real-time for accurate prediction of joint angles and moments.

One of the neural network-based methods that has gained widespread attention for sequential prediction tasks is Transformer Neural Networks (TNNs) \cite{Vaswani2017}. TNNs utilize the attention mechanism \cite{Bahdanau2015, Luong2015}, which enables these networks to effectively capture inter-sequence relationships. Originally designed for sequence translation tasks, TNNs leveraged self-attention mechanisms to identify and model the relationships between elements within a sequence, such as the words in a sentence. To address the challenges of prediction delays in real-time applications, TNNs introduced Multi-Head Attention modules. This parallel architecture significantly accelerates sequence predictions by processing multiple attention mechanisms simultaneously. For the purpose of kinematics and kinetics prediction, TNNs offer distinct advantages. They combine the general benefits of neural networks with the ability to analyze sequences of data, capturing both short-term and long-term dependencies. This feature is particularly valuable when dealing with kinematics and kinetics data, where patterns often span extended time sequences.

Additionally, the multi-head attention structure in TNNs makes them particularly suited for tasks involving fused sensory input data. This structure dynamically assigns weights based on the importance of each data source, enabling the network to prioritize certain sensor inputs over others as required by the task. This is especially beneficial in scenarios where sensory data fusion, such as IMU and sEMG signals, is essential for accurate prediction. Furthermore, TNNs' encoder-decoder architecture enhances their robustness to noisy sensor outputs, reducing the reliance on extensive preprocessing or feature extraction before feeding data into the network. TNNs were initially developed for sequence translation tasks, but their architecture can also be adapted for time series predictions with minimal modifications. This adaptability makes them a compelling choice for applications requiring real-time, sequential predictions, such as human motion modeling or robotic control.

The goal of this paper is to fuse IMU sensory data with sEMG sensory data to predict lower-limb joint kinematics and kinetics during walking, an essential activity of daily living (ADL). To validate our prediction accuracy, we will utilize data from optical motion capture cameras as the gold standard. We hypothesize that IMU data can serve as the primary source of motion kinematic information and an ancillary source of kinetic information, while sEMG data can act as the primary source of kinetic information and an ancillary source of kinematic information. By combining these two complementary data sources, we aim to achieve precise joint angle and moment predictions during walking. For this regression task, we will leverage Transformer Neural Networks (TNNs) as the model for mapping the fused sensory data to joint kinematics and kinetics. TNNs have demonstrated their ability to maintain both intra-sequence and inter-sequence relationships, making them particularly effective for capturing the complex temporal dynamics in human motion data.

Beyond standard next-step predictions, we will explore the robustness of TNNs by focusing on two prediction horizons: one near-term (30 milliseconds into the future) and one long-term (250 milliseconds into the future) across all major degrees of freedom in both lower limbs. Additionally, we will examine the network's ability to handle sudden changes in motion speed, simulating scenarios involving abrupt human movements. This investigation will evaluate the capability of our approach to adapt to real-world, dynamic conditions. We believe that the proposed trained model will provide a valuable resource for future research in this field. By utilizing a large dataset, our model can serve as a foundation for transfer learning and meta-learning applications, enabling researchers to build upon our findings for further advancements in human motion modeling and prediction.

\section{method}
\subsection{Overview of the method}
To achieve our goal, we require three types of data: IMU, sEMG, and Motion Capture. Among these, the IMU and sEMG data will be processed and used as inputs to our TNN, while the Motion Capture data will serve as ground truth for joint angles and moments rather than being directly used as input. These ground truth values represent the joint angles and moments of the hip, knee, and ankle, which the TNN is tasked with predicting. Specifically, the output of the TNN will perform multivariate regression to predict these values over two distinct horizons: a near-future prediction at 30 milliseconds and a further-future prediction at 250 milliseconds. Additionally, as part of our investigation into the network’s robustness, particularly in scenarios involving sudden changes in movement, the input data will include such variations to evaluate the network's performance under dynamic conditions. The subsequent sections detail the dataset utilized, the preprocessing steps, the preparation of data for network input, and the architecture of the proposed TNN.

\subsection{Dataset}
\label{Dataset}
To train our model for predicting joint kinematics and kinetics, it is essential to have a comprehensive dataset that includes sEMG, IMU, and optical motion capture data. For this research, we utilize an open-access dataset Camargo et al. \cite{Camargo2022}, which contains recordings from 22 healthy adult participants (average age: 21 ± 3.4 years, height: 1.7 ± 0.07 m, and weight: 68 ± 10.39 kg). The dataset includes sEMG data collected from 11 muscles in the right leg: gluteus medius, external oblique, semitendinosus, gracilis, biceps femoris, rectus femoris, vastus lateralis, vastus medialis, soleus, tibialis anterior, and gastrocnemius medialis. Raw sEMG signals were recorded at 1000 Hz and processed using a digital bandpass filter (20–400 Hz, 20th-order Butterworth). IMU data was gathered from the torso and three segments of the right lower limb—thigh, shank, and foot—with sensors placed at approximately three-quarters of the length down each segment. The IMU signals were sampled at 200 Hz and filtered with a digital low-pass filter (cutoff 100 Hz, 6th-order Butterworth).

Motion capture data were collected at 200 Hz from both legs, processed with a low-pass filter (cutoff 6 Hz, 4th-order zero-lag Butterworth), and supplemented by ground reaction force measurements at 1000 Hz, which were filtered using a low-pass filter (cutoff 15 Hz, 10th-order zero-lag Butterworth). Although joint angles were also recorded using goniometers, this data will not be used in our study. To extract joint angles and moments based on the recorded data from motion capture cameras, Opensim was used for solving the inverse kinematics and kinetics by the authors of the dataset \cite{Camargo2022}. The dataset captures a variety of locomotion modes, including level-ground walking, treadmill walking, stair climbing, and ramp traversal. For this research, we focus exclusively on treadmill walking data, as it offers an opportunity to study changes in motion patterns—one of the key objectives of our study.

In the aforementioned dataset, treadmill walking trials were conducted across 28 speeds, ranging from 0.5 m/s to 1.85 m/s in 0.05 m/s increments. Speeds were interleaved within seven trials, with four speeds per trial arranged in a sequence that simulates gradual changes in motion. For example, a trial might include transitions from rest to a slow speed, then to medium-fast and fast speeds, before slowing down to medium-slow and stopping  (e.g for the first trial speeds are 0.5 m/s, 1.2 m/s, 1.55 m/s, 0.85 m/s and trial two 0.55 m/s, 1.25 m/s, 1.6 m/s, 0.9 m/s). This pattern of incremental speed variations provides a controlled setup to investigate the effects of motion changes on our model’s predictions.

\subsection{Preprocessing}
With the dataset at hand, the next step is preprocessing the sensory data to extract meaningful information. An important aspect of preprocessing is making sure that the methods employed are suitable for real-time applications. As outlined in the dataset description in Section \ref{Dataset}, the released dataset already includes preprocessing for IMU and motion capture data, which we find adequate for our purposes. This preprocessing ensures a clean and reliable signal while maintaining real-time feasibility. However, sEMG data, being inherently noisier than IMU and motion capture data, requires additional preprocessing steps. Based on methodologies suggested by prior studies \cite{Moghadam2023, Wang2023}, the sEMG data undergoes the following procedure:

\begin{enumerate}
  \item \textit{High-pass filtering} -- A second-order Butterworth filter with a 25 Hz cutoff frequency is applied to remove low-frequency noise and baseline wander.
    \item \textit{Rectification} -- The signal is rectified to capture the absolute value, ensuring all signal values are positive.
        \item \textit{Low-pass filtering} -- A second-order Butterworth filter with a 6 Hz cutoff frequency is used to extract the signal envelope.
\end{enumerate}

This preprocessing transforms the initially processed sEMG signal into a simplified feature—the signal envelope—without relying on complex feature extraction methods. The simplicity and efficiency of this approach ensure it remains robust and applicable in real-time applications.

Since the IMU and sEMG sensors operate at different sampling frequencies (200 Hz for IMU and 1000 Hz for sEMG), a final step is required to harmonize the data for fusion. Both IMU and preprocessed sEMG data are downsampled to 100 Hz, matching the sampling frequency of the calculated joint angles and moments from motion capture data. By the end of this preprocessing stage, all data streams—IMU, processed sEMG, and joint kinematics and kinetics from motion capture cameras—are aligned and prepared for putting in the right shape for the input into the neural network.

Before feeding the data into the neural network, it is essential to structure it correctly. Our dataset includes IMU, sEMG, and ground truth joint kinematics and kinetics data from 22 participants. To ensure the generalizability of the network, we adopt a leave-one-subject-out validation and testing approach. Specifically, data from 20 participants are used for training, one participant's data is reserved for validation, and the remaining participant's data is held for testing. To prevent any learning bias toward specific trials or participants, we shuffle the data across trials and subjects prior to data preparation. This step ensures that the network learns patterns representative of the entire dataset rather than overfitting to specific individuals or sequences.

\begin{figure}[t!]
    \centering
    \includegraphics[width=1\linewidth]{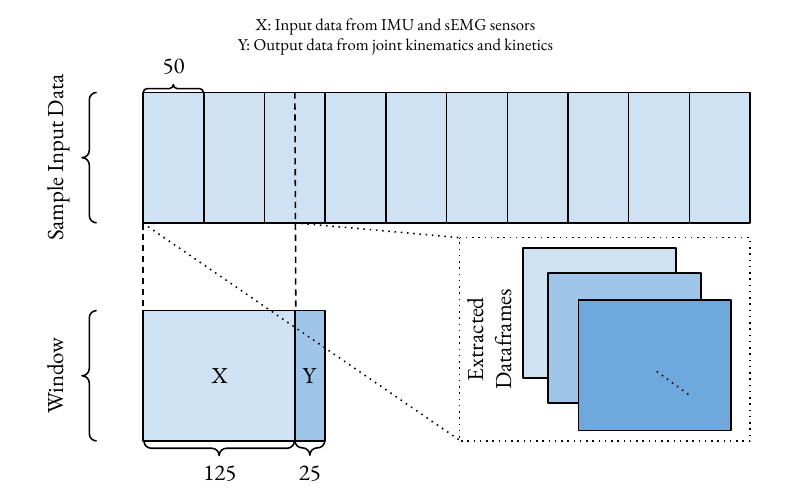}
    \caption{An illustration of data frame extraction from the input and output data.}
    \label{fig:dataframe-extraction}
\end{figure}

To prepare the data for input into the network, we employ a sliding window approach to segment it into frames, as illustrated in Fig.\ref{fig:dataframe-extraction}. Each window contains 125 samples, corresponding to 1.25 seconds at a 100 Hz sampling frequency. These 125 samples of IMU and sEMG data serve as the input, while the subsequent 25 samples (250 milliseconds) of joint angles and moments represent the target prediction for each frame. The window is then slid by four samples (0.04 seconds) to generate the next frame. To standardize the data across participants and sensors, we apply the Min-Max Normalization method. This normalization scales all features to a uniform range, facilitating consistent training and improving the stability of the learning process. By the end of this step, the dataset is segmented, normalized, and formatted into input-output pairs suitable for training, validation, and testing of the neural network.

\subsection{Neural Network Architecture}
\begin{figure}
    \vspace{0.3cm}
    \centering
    \includegraphics[width=0.8\linewidth,angle=90]{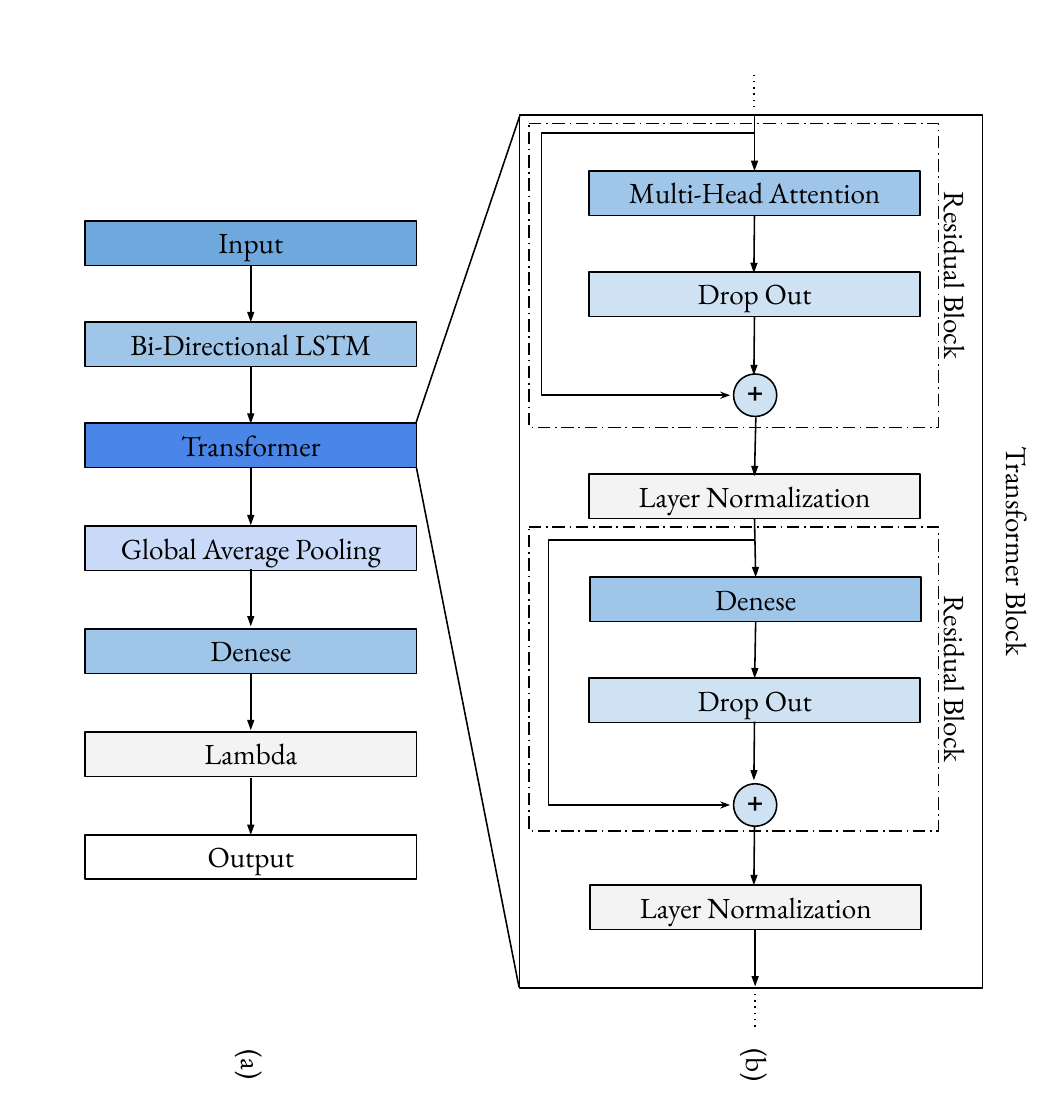}
    \caption{Neural network architecture. (a) General layers of the network, (b) Inside the transformer block.}
    \label{fig:neural-network}
\end{figure}

The proposed model for joint kinematics and kinetics prediction is a regression-based Transformer Neural Network (TNN). Since the original Transformer architecture was designed for sequence-to-sequence tasks such as sentence translation, modifications were made to adapt it for multivariable time-series prediction. The overall architecture of the network is depicted in Fig.\ref{fig:neural-network}(a). To ensure optimal weight adjustment and address issues such as gradient vanishing or explosion, we include a Bidirectional LSTM (Bi-LSTM) layer before the Transformer block. This layer processes the input sequence in both forward and backward directions, improving gradient flow and mitigating sequence forgetfulness. The processed outputs are then fed into the Transformer block, which closely follows the structure defined in \cite{Vaswani2017}. A critical component of the Transformer is its residual connections, inspired by the concept of residual blocks \cite{He2015}. As illustrated in Fig.\ref{fig:neural-network}(b), two types of residual blocks are incorporated: one encompassing the Multi-Head Attention mechanism with dropout, and the other involving a dense layer with dropout. These residual connections create bypass paths that alleviate gradient vanishing issues, enabling the network to retain longer temporal dependencies.

\begin{figure}[t!]
    \centering
    \includegraphics[width=0.6\linewidth]{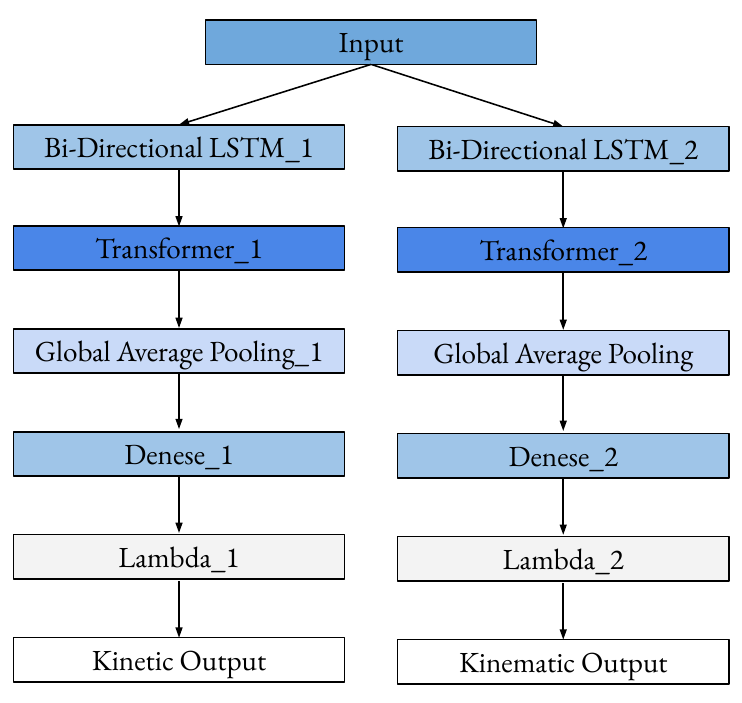}
    \caption{Proposed network for joint angles and joint moments prediction}
    \label{fig:both-networks}
\end{figure}

In the final stage of the network Fig.\ref{fig:neural-network}(a), we utilize a lambda layer to reshape the dense layer outputs into a multivariate regression format, aligning with the dimensions of the ground truth data. For this architecture, two parallel TNNs are employed, each specializing in a distinct prediction task: one for joint angles and the other for joint moments, as shown in Fig.\ref{fig:both-networks}.

The configuration of the network components is shown in Table \ref{table:hyperparameters}. All hyperparameters were selected based on extensive trial and error to optimize performance. The network was implemented and trained using the Keras and TensorFlow libraries\cite{Abadi2016}. The loss function used was Mean Squared Error (MSE), and Mean Absolute Error (MAE) was chosen as the evaluation metric. This architecture is designed to predict future joint angles and moments robustly over varying horizons, enabling accurate multivariate regression for time-series data.

\begin{table}
    \caption{Neural Network Hyperparameters}
    \centering
    \begin{tabular}{ll} \toprule
    Hyperparameter & value \\ \midrule
    Bi-LSTM units & 125 \\
    Embedding dimension & 256 \\
    Multi-Head Attention heads & 8 \\
    Dense layer neurons & 512 \\
    Dropout rate & 0.1 \\
    Layer normalization epsilon & $10^{-6}$ \\
    Optimizer-\textit{Adam},$\beta_1$ & 0.9 \\ 
    Optimizer-\textit{Adam},$\beta_2$ & 0.999 \\
    Initial learning rate & 0.0008 \\
    AMSGrad & True \\ \\\bottomrule
    \end{tabular}
    \label{table:hyperparameters}
\end{table}

\section{Results}

\begin{figure*}
    \centering
    \includegraphics[width=1\linewidth]{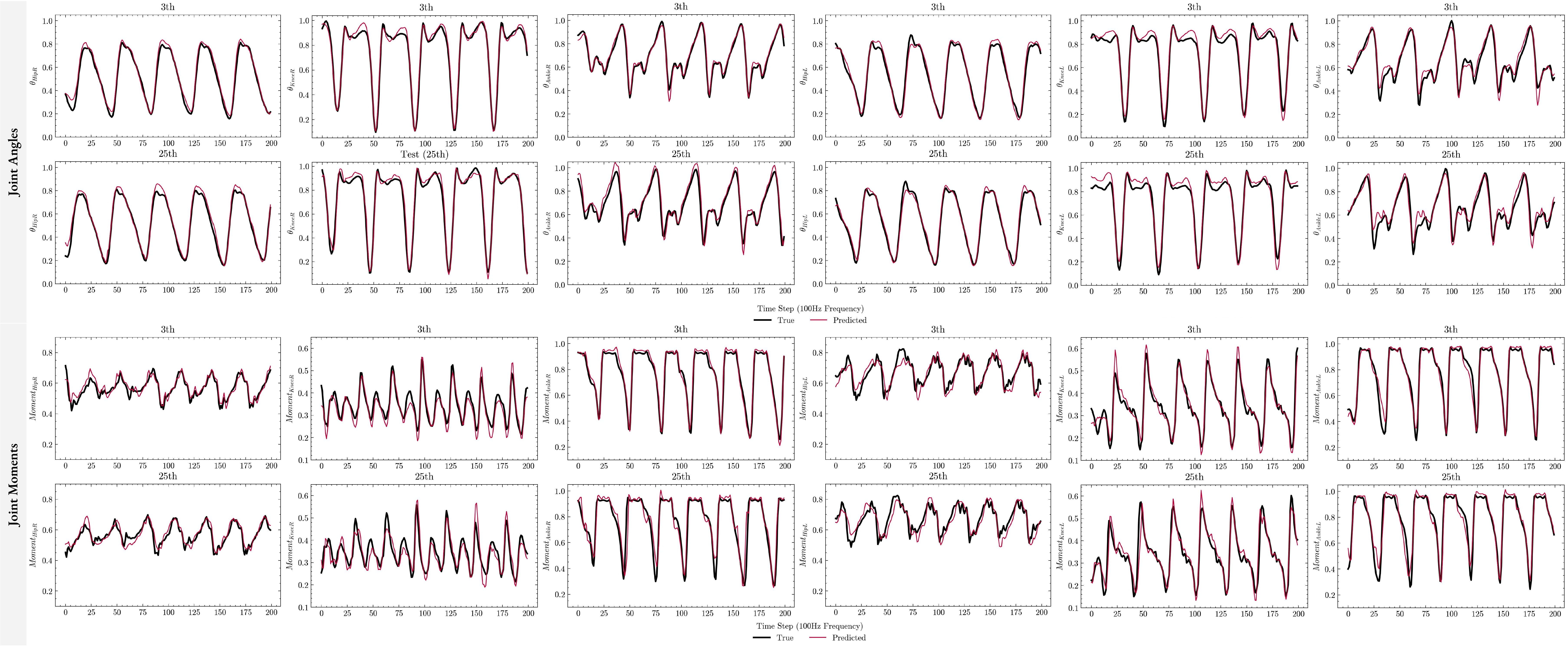}
    \caption{Full lower-limb joint angle and joint moment predictions over two horizons: 30 milliseconds (3rd value) and 250 milliseconds (25th value).}
    \label{fig:network-predictions}
\end{figure*}

The Proposed kinematic and kinetic networks were trained for 40 epochs with a batch size of 128. The primary goal of the network was to predict joint angles and moments for a 250-millisecond horizon. To assess the networks performance, two specific samples within this horizon were extracted: one at 30 milliseconds and the other at 250 milliseconds. The test data for performance evaluation came from the test subject, excluded during the preprocessing and data preparation phase. The predicted values from the network, alongside the ground truth values for the same inputs, are shown in Fig.\ref{fig:network-predictions}. To effectively evaluate the results, we have selected a random but continuous batch of 1000 data frames (equivalent to 10 seconds) from the test subject to compute errors. However, due to space constraints, only 200 data frames (2 seconds) are presented as in the visualization in Fig.\ref{fig:network-predictions}.

For a better understanding of the network performance, the prediction results are reported \textit{qualitatively} and \textit{quantitively}, over 10 seconds of data. For \textit{qualitative} evaluation, we use \textit{Coefficient of Determination (\(R^2\))}, together with \textit{Spearman correlation} (based on Shapiro-Wilk test, p-value \(<\) 0.05). These two statistical metrics can help us to understand the similarity and compatibility of the prediction curves and ground truth curves that we see in Fig.\ref{fig:network-predictions}. For \textit{quantitive} evaluation, we report, Mean Absolute Error (MAE), Root Mean Squared Error (RMSE), and normalized Root Mean Squared Error (nRMSE). All of the aforementioned metrics are evaluated both for Close Horizon (CH), and Distant Horizon (DH) and reported in the Table \ref{tab:results}.

\begin{table*}[]
\centering
\caption{Spearman's \(\rho\), coefficient of determination \(R^2\), Mean Absolute Error (MAE), Maximum MAE, Root Mean Squared Error (RMSE), and normalized Root Mean Squared Error (nRMSE) for predicted values by the network in close horizon (30 millisecond) - \textit{CH}, and distant horizon (250 millisecond) - \textit{DH} over a 10-second period. Both joint kinematics and joint kinetics are included in the table. MAE, RMSE, and nRMSE values are reported in percentage.}
\label{tab:results}
\resizebox{\textwidth}{!}{
\begin{tabular}{cc|ll|cc|cc|cc|cc|}
\cline{3-12}
\multicolumn{1}{l}{}                              & \multicolumn{1}{l|}{} & \multicolumn{2}{c|}{\textbf{\(\rho\)}}                              & \multicolumn{2}{c|}{\textbf{\(R^2\)}} & \multicolumn{2}{c|}{\textbf{MAE*}}              & \multicolumn{2}{c|}{\textbf{RMSE*}}             & \multicolumn{2}{c|}{\textbf{nRMSE*}}   \\ \cline{2-12} 
\multicolumn{1}{l|}{}                             & \textbf{Joint}        & \multicolumn{1}{c|}{\textbf{CH}} & \multicolumn{1}{c|}{\textbf{DH}} & \multicolumn{1}{c|}{\textbf{CH}}   & \textbf{DH}   & \multicolumn{1}{c|}{\textbf{CH}} & \textbf{DH} & \multicolumn{1}{c|}{\textbf{CH}} & \textbf{DH} & \multicolumn{1}{c|}{\textbf{CH}} & \textbf{DH} \\ \hline
\multicolumn{1}{|c|}{\multirow{6}{*}{Kinematics}} & Hip Right             & \multicolumn{1}{l|}{0.995}            & 0.992                                  & \multicolumn{1}{c|}{0.988}              & 0.983              & \multicolumn{1}{c|}{2.59}            &  3.04           & \multicolumn{1}{c|}{3.1}            & 3.7           & \multicolumn{1}{c|}{3.84}            & 4.56    \\ \cline{2-12} 
\multicolumn{1}{|c|}{}                            & Knee Right            & \multicolumn{1}{l|}{0.982}            & 0.974                                  & \multicolumn{1}{c|}{0.989}              & 0.980              & \multicolumn{1}{c|}{2.64}            & 3.16            & \multicolumn{1}{c|}{3.2}            & 3.9            & \multicolumn{1}{c|}{3.36}            & 4.14    \\ \cline{2-12} 
\multicolumn{1}{|c|}{}                            & Ankle Right           & \multicolumn{1}{l|}{0.985}            & 0.976                                  & \multicolumn{1}{c|}{0.971}              & 0.954              & \multicolumn{1}{c|}{2.15}            & 3.47            & \multicolumn{1}{c|}{2.6}            & 4.4            & \multicolumn{1}{c|}{3.08}            & 4.84   \\ \cline{2-12} 
\multicolumn{1}{|c|}{}                            & Hip Left              & \multicolumn{1}{l|}{0.993}            & 0.992                                  & \multicolumn{1}{c|}{0.986}              & 0.988              & \multicolumn{1}{c|}{2.97}            & 2.53           & \multicolumn{1}{c|}{3.5}            & 3.2            & \multicolumn{1}{c|}{4.16}            & 3.81   \\ \cline{2-12} 
\multicolumn{1}{|c|}{}                            & Knee Left             & \multicolumn{1}{l|}{0.978}            & 0.972                                  & \multicolumn{1}{c|}{0.979}              & 0.972              & \multicolumn{1}{c|}{3.48}            & 4.04            & \multicolumn{1}{c|}{4.3}            & 5            & \multicolumn{1}{c|}{4.61}            & 5.61   \\ \cline{2-12} 
\multicolumn{1}{|c|}{}                            & Ankle Left            & \multicolumn{1}{l|}{0.973}            & 0.963                                  & \multicolumn{1}{c|}{0.963}              & 0.949              & \multicolumn{1}{c|}{2.86}            & 3.58            & \multicolumn{1}{c|}{4.2}            & 5           & \multicolumn{1}{c|}{4.62}            & 5.59   \\ \hline
\multicolumn{1}{|c|}{\multirow{6}{*}{Kinetics}}   & Hip Right             & \multicolumn{1}{c|}{0.985}            & 0.979                                  & \multicolumn{1}{c|}{0.955}              & 0.933              & \multicolumn{1}{c|}{2.14}            & 2.50            & \multicolumn{1}{c|}{2.7}            & 3.1           & \multicolumn{1}{c|}{4.54}            & 5.35    \\ \cline{2-12} 
\multicolumn{1}{|c|}{}                            & Knee Right            & \multicolumn{1}{c|}{0.983}            & 0.966                                  & \multicolumn{1}{c|}{0.959}              & 0.924              & \multicolumn{1}{c|}{2.15}            & 2.5            & \multicolumn{1}{c|}{2.6}            & 3.3            & \multicolumn{1}{c|}{4.28}            & 5.35  \\ \cline{2-12} 
\multicolumn{1}{|c|}{}                            & Ankle Right           & \multicolumn{1}{c|}{0.966}            & 0.937                                  & \multicolumn{1}{c|}{0.981}              & 0.945              & \multicolumn{1}{c|}{2.40}            & 4.1            & \multicolumn{1}{c|}{3.1}            & 5.9            & \multicolumn{1}{c|}{3.46}            & 6.97   \\ \cline{2-12} 
\multicolumn{1}{|c|}{}                            & Hip Left              & \multicolumn{1}{l|}{0.981}            & 0.979                                  & \multicolumn{1}{c|}{0.965}              & 0.952              & \multicolumn{1}{c|}{2.5}            & 2.7             & \multicolumn{1}{c|}{3.2}            & 3.5            & \multicolumn{1}{c|}{4.79}            & 5.38  \\ \cline{2-12} 
\multicolumn{1}{|c|}{}                            & Knee Left             & \multicolumn{1}{l|}{0.966}            & 0.956                                  & \multicolumn{1}{c|}{0.952}              & 0.930              & \multicolumn{1}{c|}{2.2}            & 2.6            & \multicolumn{1}{c|}{3.0}            & 3.8           & \multicolumn{1}{c|}{4.14}            & 5.38   \\ \cline{2-12} 
\multicolumn{1}{|c|}{}                            & Ankle Left            & \multicolumn{1}{l|}{0.952}            & 0.937                                  & \multicolumn{1}{c|}{0.962}              & 0.950              & \multicolumn{1}{c|}{3.8}            & 4.1            & \multicolumn{1}{c|}{5.1}            & 6.1            & \multicolumn{1}{c|}{5.46}            & 6.78   \\ \hline
\end{tabular}%
}
\vspace{1ex}

\raggedright * The reported values of MAE, RMSE, nRMSE in the table are multiplied by 100. \par

\end{table*}

\section{Discussion}

The primary objective of this study was to predict full lower-limb joint angles and joint moments over extended temporal horizons. To this end, we employed data from IMU and sEMG sensors as inputs to two separately designed Transformer Neural Networks (TNNs): one dedicated to joint angle prediction and the other to joint moment prediction. Two distinct prediction horizons were defined: a short-term horizon of 30 milliseconds (close-horizon, CH) and a long-term horizon of 250 milliseconds (distant-horizon, DH). These horizons were selected to systematically evaluate the networks’ predictive performance across both immediate and extended future timeframes. Our underlying hypothesis posits that the combined use of IMU and sEMG signals offers complementary kinematic and neuromuscular information, which is critical for accurately forecasting both joint angles and joint moments. 

Both Fig. \ref{fig:network-predictions} and the statistical results from Table \ref{tab:results} demonstrate that the network accurately predicted kinematics and kinetics for the unseen test subject. The Spearman correlation test for kinematic predictions indicates a statistically significant and strong positive correlation between the predicted values and ground truth data. Specifically, for the close-horizon (CH) prediction, the lowest value of the correlation coefficient was \(\rho\) = 0.973 (\(p\) \(<\) 0.001), while for the distant-horizon (DH) prediction, it was \(\rho\) = 0.963 (\(p\) \(<\) 0.001). Among all joints, the left ankle joint had the lowest correlation coefficient, yet still exhibited a strong positive relationship.

In addition to Spearman's correlation coefficient, the predictive performance of the network was further evaluated using the coefficient of determination \(R^2\). The lowest \(R^2\) value achieved by the model was 0.963 for CH and 0.949 for DH in the left ankle joint, with an RMSE of 0.042\degree and an MAE of 0.028\degree in CH, and an RMSE of 0.050\degree with an MAE of 0.035\degree in DH. These results indicate that approximately 95\% (CH) and 92\% (DH) of the variance in the ground truth joint angles for the least accurate joint can still be explained by the model’s predictions. This suggests that the proposed network exhibits strong predictive capabilities for lower-limb kinematics and kinetics.

For the predicted kinetic results, the Spearman correlation test indicates a statistically significant correlation between the predicted joint moments and the ground truth joint moments. The lowest Spearman correlation value in CH is \(\rho\) = 0.973 (\(p\) \(<\) 0.001), and in DH, equals to  \(\rho\) = 0.973 (\(p\) \(<\) 0.001) for the left ankle joint. The \(R^2\) score across all joint moments suggests a strong predictive ability. The lowest \(R^2\) score for predicted joint moments in CH belongs to the left knee, with a score of 0.952 (MAE = 0.022 N.m/kg and RMSE = 0.03 N.m/kg), while in DH, the lowest \(R^2\) score belongs to the right knee, with a score of 0.924 (MAE = 0.025 N.m/kg and RMSE = 0.033 N.m/kg). Meanwhile, the highest recorded MAE and RMSE for joint moments occur at the left ankle in both CH (MAE = 0.038 N.m/kg, RMSE = 0.051 N.m/kg) and DH (MAE = 0.041 N.m/kg, RMSE = 0.061 N.m/kg). The highest joint moment prediction errors based on normalized RMSE are observed at the left ankle in CH, with 5.46\% error, and at the right ankle in DH, with 6.97\% error.

In joint kinematics, based on MAE and RMSE, the joint with the highest error in both CH (MAE = 0.048\degree, RMSE = 0.043\degree) and DH (MAE = 0.0404\degree, RMSE = 0.050\degree) is the left knee. However, the RMSE of the left ankle in DH has the same value as the left knee. According to normalized RMSE, the highest error in CH is for the left ankle (4.62\%), while in DH, the highest error is for the left knee (5.61\%). As observed, the higher error values correspond to the leg without attached sensors, suggesting that while it is possible to predict joint angles and moments for one leg using sensory data from the opposite leg, achieving equally accurate predictions requires sensor attachment on both legs, which aligns with expectations. Additionally, the joint with the highest prediction error among our six joints is typically the left ankle, indicating that more sensory information is needed for improved accuracy in left ankle predictions.

Compared to similar research, the closest study to our work is that of Mohammadi-Moghadam et al. \cite{Moghadam2023}. By extracting key features from fused sEMG and IMU signals and leveraging a CNN as their predictor model, they successfully predicted joint angles and joint moments across three planes of motion. However, a limitation of their work is that their network is constrained to real-world scenarios but does not support real-time applications. Moreover, their prediction horizon is limited to one-step-ahead forecasting, which may not be suitable for robotic control systems. Our proposed method, on the other hand, addresses these limitations by enabling real-time predictions and extending the prediction horizon, making it more applicable for robotic control. Additionally, our approach improves both joint angle and joint moment predictions, outperforming their best-reported \textit{inter-subject} RMSE values.

For the hip joint angle, our RMSE results are 0.031$^{\circ}$ and 0.037$^{\circ}$ for the right leg in CH and DH predictions, and 0.035$^{\circ}$ and 0.032$^{\circ}$ for the left leg in CH and DH, respectively. These values indicate a significant improvement compared to their reported value of 4.79$^{\circ}$. For the knee joint angle, our RMSE values are 0.032$^{\circ}$ and 0.039$^{\circ}$ for the right leg in CH and DH, respectively, and 0.043$^{\circ}$ and 0.050$^{\circ}$ for the left leg in CH and DH, respectively. Similar to our hip joint angle results, these values demonstrate an improvement compared to their reported value of 5.46$^{\circ}$. For the ankle joint angle, our RMSE values are 0.026$^{\circ}$ and 0.044$^{\circ}$ for the right leg in CH and DH, respectively, and 0.042$^{\circ}$ and 0.050$^{\circ}$ for the left leg in CH and DH, respectively. Meanwhile, their reported RMSE for the ankle joint angle was 6.52$^{\circ}$. These results indicate that our proposed model for kinematic prediction outperformed their proposed model across all three joint angles.

For the hip joint moment, our network achieved an RMSE of 0.021 N.m/kg for CH and 0.025 N.m/kg for DH of the right leg. For the left hip joint moment, we obtained an RMSE of 0.025 N.m/kg for CH and 0.027 N.m/kg for DH. The reported inter-subject hip moment RMSE for the baseline research \cite{Moghadam2023} is not explicitly mentioned in the paper. However, based on the boxplot, the lower limit of the hip flexion/extension error median is approximately 0.15 N.m/kg, which corresponds to the Random Forest method. For the right knee joint moment, our proposed method achieved RMSE values of 0.026 N.m/kg and 0.033 N.m/kg for CH and DH, respectively, while for the left knee joint moment, the values were 0.030 N.m/kg and 0.038 N.m/kg. In contrast, Mohammadi-Moghadam et al. reported an inter-subject RMSE of 0.187 N.m/kg for the knee joint moment. For the ankle joint moment, the right leg RMSE was 0.031 N.m/kg for CH and 0.059 N.m/kg for DH, while for the left leg, the values were 0.051 N.m/kg and 0.061 N.m/kg, respectively. The best-reported RMSE value from \cite{Moghadam2023} was 0.066 N.m/kg for \textit{intra-subject} prediction (which is generally more accurate than \textit{inter-subject} RMSE, as some portion of the user's data is seen by the network). As a result, we can conclude that our model outperformed the baseline model in kinetic prediction as well.

Wang et al. \cite{Wang2025} recently proposed Dual Transformer Network (DTN) to solve joint kinematics and kinetics using sEMG signals. Compared to the proposed DTN method, our model extends beyond short-term prediction by forecasting joint trajectories multiple steps into the future, enabling anticipatory control strategies that are particularly beneficial for wearable assistive devices. In contrast, the DTN model is limited to a fixed 50 ms prediction horizon, which may not be sufficient for preemptive adjustments in dynamic environments. Additionally, our framework is designed to predict the full set of lower-limb joint kinematics and kinetics—including hip, knee, and ankle—across both legs. This holistic modeling approach provides a richer understanding of bilateral gait dynamics and supports applications in both rehabilitation and robotic assistance. The DTN model, by contrast, is constrained to uniplanar predictions of one leg, limiting its versatility in real-world use cases. 

Furthermore, in compare to Wang et al. \cite{Wang2025} proposed model, our model demonstrates lower root-mean-square error (RMSE) across all predicted joint angles with DTN's reported RMSE values being an order of magnitude higher. This highlights the superior numerical accuracy of our model, which is particularly critical in applications involving real-time control or biomechanical assessment. However, it is noteworthy that DTN reports higher \(R^2\) and Spearman’s correlation coefficients, indicating stronger alignment in the temporal patterns and rank-order consistency of predicted signals relative to ground truth. This suggests that while DTN effectively captures the shape and trend of joint trajectories, it lacks precision in magnitude, potentially limiting its use in applications requiring fine-grained kinematic accuracy. Although our kinematic model outperforms their model, the predicted joint moments in our model in the right leg for both CH and DH have slightly better accuracy in the the hip and ankle joints, with their reported values being 0.048 Nm/Kg and 0.060 Nm/Kg, respectively. While, knee moment is the same for CH and slightly increased error for DH, with their reported value for knee moment being 0.026 Nm/Kg. For the left leg, hip and knee joint in both CH and DH have lower RMSE values in our model, while the ankle joint has a lower RMSE in CH and a slightly higher RMSE in DH. Moreover, our model benefits from multi-modal sensor fusion by integrating both sEMG and IMU data, whereas DTN relies solely on sEMG, which can be sensitive to signal variability and lacks motion-state context. Taken together, our approach offers a more robust and precise solution for joint angle and moment prediction, especially in scenarios where absolute accuracy is essential.

\section{Conclusion}
Our goal in this research was to predict full lower-limb joint moments and joint angles over long horizons based on sensory data. To achieve this, we developed a regression network inspired by Transformer Neural Networks (TNNs) and trained the model using IMU and sEMG data. Our hypothesis was that IMU and sEMG signals serve as complementary sources of information for both kinematic and kinetic predictions. The results indicate that our network can accurately predict both joint moments and joint angles, demonstrating superior performance compared to the baseline model. Furthermore, we believe the proposed model can be effectively deployed in real-time applications, as its parallel neural network structure enhances computational efficiency, making it well-suited for real-time scenarios.

\section{Future Works}
As a potential direction for future research, one could explore ways to enhance the computational efficiency of the network by optimizing the input data. Specifically, investigating whether accurate predictions can be achieved using a reduced number of inputs from IMU signals or sEMG signals could be beneficial. Additionally, in this study, sEMG and IMU signals were treated equally for both the joint angle prediction network and the joint moment prediction network. However, introducing a dynamic weighting mechanism between IMU and sEMG inputs, based on the specific prediction task (whether kinematic or kinetic), may further improve prediction accuracy.

\bibliographystyle{ieeetr}

\begin{thebibliography}{10}

\bibitem{Buchanan2005}
T.~S. Buchanan, D.~G. Lloyd, K.~Manal, and T.~F. Besier, ``Estimation of muscle forces and joint moments using a forward-inverse dynamics model,'' in {\em Medicine and Science in Sports and Exercise}, vol.~37, pp.~1911--1916, 11 2005.

\bibitem{Crenna2011}
P.~Crenna and C.~Frigo, ``Dynamics of the ankle joint analyzed through moment-angle loops during human walking: Gender and age effects,'' {\em Human Movement Science}, vol.~30, pp.~1185--1198, 12 2011.

\bibitem{Chowdhury2013}
S.~Chowdhury and N.~Kumar, ``Estimation of forces and moments of lower limb joints from kinematics data and inertial properties of the body by using inverse dynamics technique,'' 2013.

\bibitem{Hov1990}
M.~G. Hoy, F.~E. Zajac, and M.~E. Gordon, ``A musculoskeletal model of the human lower extremity: The effect of muscle, tendon, and moment arm on the moment-angle relationship of musculotendon actuators at the hip, knee, and ankle,'' {\em Journal of biomechanics}, 1990.

\bibitem{Maganaris2004}
C.~N. Maganaris, ``A predictive model of moment-angle characteristics in human skeletal muscle: Application and validation in muscles across the ankle joint,'' {\em Journal of Theoretical Biology}, vol.~230, pp.~89--98, 9 2004.

\bibitem{Ceseracciu2014}
E.~Ceseracciu, Z.~Sawacha, and C.~Cobelli, ``Comparison of markerless and marker-based motion capture technologies through simultaneous data collection during gait: Proof of concept,'' {\em PLoS ONE}, vol.~9, 3 2014.

\bibitem{Kanko2021}
R.~M. Kanko, E.~K. Laende, E.~M. Davis, W.~S. Selbie, and K.~J. Deluzio, ``Concurrent assessment of gait kinematics using marker-based and markerless motion capture,'' {\em Journal of Biomechanics}, vol.~127, 10 2021.

\bibitem{Thalmann1997}
D.~Thalmann and M.~Panne, {\em Computer Animation and Simulation ’97}, vol.~VIII.
\newblock Springer Vienna, 1~ed., 9 1997.

\bibitem{Herda2000}
L.~Herda, P.~Fua, R.~Plankers, R.~Boulic, and D.~Thalmann, ``Skeleton-based motion capture for robust reconstruction of human motion,'' {\em IEEE}, 5 2000.

\bibitem{Liu2006}
G.~Liu and L.~McMillan, ``Estimation of missing markers in human motion capture,'' {\em Visual Computer}, vol.~22, pp.~721--728, 9 2006.

\bibitem{Aristidou2008}
A.~Aristidou, J.~Cameron, and J.~Lasenby, ``Real-time estimation of missing markers in human motion capture-aristidou,'' {\em IEEE}, 5 2008.

\bibitem{Aristidou2013}
A.~Aristidou and J.~Lasenby, ``Real-time marker prediction and cor estimation in optical motion capture,'' {\em Visual Computer}, vol.~29, pp.~7--26, 2013.

\bibitem{Buchanan2004}
T.~S. Buchanan, D.~G. Lloyd, K.~Manal, and T.~F. Besier, ``Neuromusculoskeletal modeling: Estimation of muscle forces and joint moments and movements from measurements of neural command,'' 2004.

\bibitem{Gabbasov2015}
B.~Gabbasov, I.~Danilov, I.~M. Afanasyev, and E.~Magid, ``Toward a human-like biped robot gait: Biomechanical analysis of human locomotion recorded by kinect-based motion capture system,'' {\em IEEE International Symposium on Mechatronics and its Applications}, 12 2015.

\bibitem{Wang2021}
J.~ping Wang, S.~hua Wang, Y.~qing Wang, H.~Hu, J.~wei Yu, X.~Zhao, J.~lai Liu, X.~Chen, and Y.~Li, ``A data process of human knee joint kinematics obtained by motion-capture measurement,'' {\em BMC Medical Informatics and Decision Making}, vol.~21, 12 2021.

\bibitem{Willemsen1990}
A.~Willemsen, J.~V. Alste, and H.~Boom, ``Real-time gait assessment utilizing a new way of accelerometry.''

\bibitem{MoeNilssen2004}
R.~Moe-Nilssen and J.~L. Helbostad, ``Estimation of gait cycle characteristics by trunk accelerometry,'' {\em Journal of Biomechanics}, vol.~37, pp.~121--126, 2004.

\bibitem{Selles2005}
R.~W. Selles, M.~A. Formanoy, J.~B. Bussmann, P.~J. Janssens, and H.~J. Stam, ``Automated estimation of initial and terminal contact timing using accelerometers; development and validation in transtibial amputees and controls,'' {\em IEEE Transactions on Neural Systems and Rehabilitation Engineering}, vol.~13, pp.~81--88, 3 2005.

\bibitem{Picerno2017}
P.~Picerno, ``25 years of lower limb joint kinematics by using inertial and magnetic sensors: A review of methodological approaches,'' 1 2017.

\bibitem{Goulermas2005}
J.~Y. Goulermas, D.~Howard, C.~J. Nester, R.~K. Jones, and L.~Ren, ``Regression techniques for the prediction of lower limb kinematics,'' {\em Journal of Biomechanical Engineering}, vol.~127, pp.~1020--1024, 11 2005.

\bibitem{Findlow2008}
A.~Findlow, J.~Y. Goulermas, C.~Nester, D.~Howard, and L.~P. Kenney, ``Predicting lower limb joint kinematics using wearable motion sensors,'' {\em Gait and Posture}, vol.~28, pp.~120--126, 7 2008.

\bibitem{Lim2020}
H.~Lim, B.~Kim, and S.~Park, ``Prediction of lower limb kinetics and kinematics during walking by a single imu on the lower back using machine learning,'' {\em Sensors (Switzerland)}, vol.~20, 1 2020.

\bibitem{Weygers2020}
I.~Weygers, M.~Kok, M.~Konings, H.~Hallez, H.~D. Vroey, and K.~Claeys, ``Inertial sensor-based lower limb joint kinematics: A methodological systematic review,'' 2 2020.

\bibitem{Niswander2020}
W.~Niswander, W.~Wang, and K.~Kontson, ``Optimization of imu sensor placement for the measurement of lower limb joint kinematics,'' {\em Sensors (Switzerland)}, vol.~20, pp.~1--16, 11 2020.

\bibitem{Pitt2020}
W.~Pitt, S.~H. Chen, and L.~S. Chou, ``Using imu-based kinematic markers to monitor dual-task gait balance control recovery in acutely concussed individuals,'' {\em Clinical Biomechanics}, vol.~80, 12 2020.

\bibitem{Majumder2021}
S.~Majumder and M.~J. Deen, ``Wearable imu-based system for real-time monitoring of lower-limb joints,'' {\em IEEE Sensors Journal}, vol.~21, pp.~8267--8275, 3 2021.

\bibitem{McGrath2022}
T.~McGrath and L.~Stirling, ``Body-worn imu-based human hip and knee kinematics estimation during treadmill walking,'' {\em Sensors}, vol.~22, 4 2022.

\bibitem{Carcreff2022}
L.~Carcreff, G.~Payen, G.~Grouvel, F.~Massé, and S.~Armand, ``Three-dimensional lower-limb kinematics from accelerometers and gyroscopes with simple and minimal functional calibration tasks: Validation on asymptomatic participants,'' {\em Sensors}, vol.~22, 8 2022.

\bibitem{Versteyhe2020}
M.~Versteyhe, H.~D. Vroey, F.~Debrouwere, H.~Hallez, and K.~Claeys, ``A novel method to estimate the full knee joint kinematics using low cost imu sensors for easy to implement low cost diagnostics,'' {\em Sensors (Switzerland)}, vol.~20, 3 2020.

\bibitem{LoraMillan2021}
J.~S. Lora-Millan, A.~F. Hidalgo, and E.~Rocon, ``An imus-based extended kalman filter to estimate gait lower limb sagittal kinematics for the control of wearable robotic devices,'' {\em IEEE Access}, vol.~9, pp.~144550--144554, 2021.

\bibitem{Sy2021}
L.~Sy, M.~Raitor, M.~D. Rosario, H.~Khamis, L.~Kark, N.~H. Lovell, and S.~J. Redmond, ``Estimating lower limb kinematics using a reduced wearable sensor count,'' {\em IEEE Transactions on Biomedical Engineering}, vol.~68, pp.~1293--1304, 4 2021.

\bibitem{Potter2022}
M.~V. Potter, S.~M. Cain, L.~V. Ojeda, R.~D. Gurchiek, R.~S. McGinnis, and N.~C. Perkins, ``Evaluation of error-state kalman filter method for estimating human lower-limb kinematics during various walking gaits,'' {\em Sensors}, vol.~22, 11 2022.

\bibitem{Lee2020}
M.~Lee and S.~Park, ``Estimation of three-dimensional lower limb kinetics data during walking using machine learning from a single imu attached to the sacrum,'' {\em Sensors (Switzerland)}, vol.~20, pp.~1--16, 11 2020.

\bibitem{Lee2022}
C.~J. Lee and J.~K. Lee, ``Inertial motion capture-based wearable systems for estimation of joint kinetics: A systematic review,'' 4 2022.

\bibitem{Krishnakumar2024}
S.~Krishnakumar, B.~J.~F. van Beijnum, C.~T. Baten, P.~H. Veltink, and J.~H. Buurke, ``Estimation of kinetics using imus to monitor and aid in clinical decision-making during acl rehabilitation: A systematic review,'' 4 2024.

\bibitem{Dorschky2019}
E.~Dorschky, M.~Nitschke, A.~K. Seifer, A.~J. van~den Bogert, and B.~M. Eskofier, ``Estimation of gait kinematics and kinetics from inertial sensor data using optimal control of musculoskeletal models,'' {\em Journal of Biomechanics}, vol.~95, 10 2019.

\bibitem{Mundt2020a}
M.~Mundt, W.~Thomsen, T.~Witter, A.~Koeppe, S.~David, F.~Bamer, W.~Potthast, and B.~Markert, ``Prediction of lower limb joint angles and moments during gait using artificial neural networks,'' {\em Medical and Biological Engineering and Computing}, vol.~58, pp.~211--225, 2020.

\bibitem{Mundt2020b}
M.~Mundt, A.~Koeppe, S.~David, T.~Witter, F.~Bamer, W.~Potthast, and B.~Markert, ``Estimation of gait mechanics based on simulated and measured imu data using an artificial neural network,'' {\em Frontiers in Bioengineering and Biotechnology}, vol.~8, pp.~1--16, 2020.

\bibitem{Stanev2021}
D.~Stanev, K.~Filip, D.~Bitzas, S.~Zouras, G.~Giarmatzis, D.~Tsaopoulos, and K.~Moustakas, ``Real-time musculoskeletal kinematics and dynamics analysis using marker-and imu-based solutions in rehabilitation,'' {\em Sensors}, vol.~21, pp.~1--20, 3 2021.

\bibitem{Yi2021}
C.~Yi, F.~Jiang, M.~Z.~A. Bhuiyan, C.~Yang, X.~Gao, H.~Guo, J.~Ma, and S.~Su, ``Smart healthcare-oriented online prediction of lower-limb kinematics and kinetics based on data-driven neural signal decoding,'' {\em Future Generation Computer Systems}, vol.~114, pp.~96--105, 1 2021.

\bibitem{Moghadam2023}
S.~M. Moghadam, T.~Yeung, and J.~Choisne, ``A comparison of machine learning models’ accuracy in predicting lower-limb joints’ kinematics, kinetics, and muscle forces from wearable sensors,'' {\em Scientific Reports}, vol.~13, pp.~1--16, 2023.

\bibitem{Dimitrov2023}
H.~Dimitrov, A.~M. Bull, and D.~Farina, ``High-density emg, imu, kinetic, and kinematic open-source data for comprehensive locomotion activities,'' {\em Scientific Data}, vol.~10, 12 2023.

\bibitem{Ardestani2014}
M.~M. Ardestani, X.~Zhang, L.~Wang, Q.~Lian, Y.~Liu, J.~He, D.~Li, and Z.~Jin, ``Human lower extremity joint moment prediction: A wavelet neural network approach,'' {\em Expert Systems with Applications}, vol.~41, pp.~4422--4433, 7 2014.

\bibitem{Brantley2017}
J.~A. Brantley, S.~Member, T.~P. Luu, S.~Nakagome, J.~L. Contreras-Vidal, and S.~Member, ``Prediction of lower-limb joint kinematics from surface emg during overground locomotion,'' 2017.

\bibitem{Chen2018}
J.~Chen, X.~Zhang, Y.~Cheng, and N.~Xi, ``Surface emg based continuous estimation of human lower limb joint angles by using deep belief networks,'' {\em Biomedical Signal Processing and Control}, vol.~40, pp.~335--342, 2 2018.

\bibitem{Bi2019}
L.~Bi, A.~Feleke, and C.~Guan, ``A review on emg-based motor intention prediction of continuous human upper limb motion for human-robot collaboration,'' 5 2019.

\bibitem{Xiong2019}
B.~Xiong, N.~Zeng, H.~Li, Y.~Yang, Y.~Li, M.~Huang, W.~Shi, M.~Du, and Y.~Zhang, ``Intelligent prediction of human lower extremity joint moment: An artificial neural network approach,'' {\em IEEE Access}, vol.~7, pp.~29973--29980, 2019.

\bibitem{Huang2019}
Y.~Huang, Z.~He, Y.~Liu, R.~Yang, X.~Zhang, G.~Cheng, J.~Yi, J.~P. Ferreira, and T.~Liu, ``Real-time intended knee joint motion prediction by deep-recurrent neural networks,'' {\em IEEE Sensors Journal}, vol.~19, pp.~11503--11509, 12 2019.

\bibitem{Liang2021}
J.~Liang, Z.~Shi, F.~Zhu, W.~Chen, X.~Chen, and Y.~Li, ``Gaussian process autoregression for joint angle prediction based on semg signals,'' {\em Frontiers in Public Health}, vol.~9, 5 2021.

\bibitem{Sitole2023}
S.~P. Sitole and F.~C. Sup, ``Continuous prediction of human joint mechanics using emg signals: A review of model-based and model-free approaches,'' {\em IEEE Transactions on Medical Robotics and Bionics}, vol.~5, pp.~528--546, 8 2023.

\bibitem{Yaakoubi2023}
N.~A.~E. Yaakoubi, C.~McDonald, and O.~Lennon, ``Prediction of gait kinematics and kinetics: A systematic review of emg and eeg signal use and their contribution to prediction accuracy,'' 10 2023.

\bibitem{Olney1985}
S.~J. Olney and D.~A. Winter, ``Predictions of knee and ankle moments of force in walking from emg and kinematic data,'' 1985.

\bibitem{Koo2005}
T.~K. Koo and A.~F. Mak, ``Feasibility of using emg driven neuromusculoskeletal model for prediction of dynamic movement of the elbow,'' {\em Journal of Electromyography and Kinesiology}, vol.~15, pp.~12--26, 2 2005.

\bibitem{Shao2009}
Q.~Shao, D.~N. Bassett, K.~Manal, and T.~S. Buchanan, ``An emg-driven model to estimate muscle forces and joint moments in stroke patients,'' {\em Computers in Biology and Medicine}, vol.~39, pp.~1083--1088, 12 2009.

\bibitem{Camargo2022}
J.~Camargo, D.~Molinaro, and A.~Young, ``Predicting biological joint moment during multiple ambulation tasks,'' {\em Journal of Biomechanics}, vol.~134, 3 2022.

\bibitem{Rabe2022}
K.~G. Rabe and N.~P. Fey, ``Evaluating electromyography and sonomyography sensor fusion to estimate lower-limb kinematics using gaussian process regression,'' {\em Frontiers in Robotics and AI}, vol.~9, 3 2022.

\bibitem{Koike1995}
Y.~Koike and M.~Kawato, ``Biological cybernetics estimation of dynamic joint torques and trajectory formation from surface electromyography signals using a neural network model,'' {\em Biol. Cybern}, vol.~73, p.~300, 1995.

\bibitem{Lloyd2003}
D.~G. Lloyd and T.~F. Besier, ``An emg-driven musculoskeletal model to estimate muscle forces and knee joint moments in vivo,'' {\em Journal of Biomechanics}, vol.~36, pp.~765--776, 6 2003.

\bibitem{Zhang2013}
Q.~Zhang, R.~Hosoda, and G.~Venture, ``Human joint motion estimation for electromyography (emg)-based dynamic motion control,'' {\em 35th Annual International Conference of the IEEE EMBS}, 2013.

\bibitem{Merletti2020}
R.~Merletti and G.~L. Cerone, ``Tutorial. surface emg detection, conditioning and pre-processing: Best practices,'' {\em Journal of Electromyography and Kinesiology}, vol.~54, 10 2020.

\bibitem{Baghdadi2018}
A.~Baghdadi, L.~A. Cavuoto, and J.~L. Crassidis, ``Hip and trunk kinematics estimation in gait through kalman filter using imu data at the ankle,'' {\em IEEE Sensors Journal}, vol.~18, pp.~4253--4260, 5 2018.

\bibitem{Abdelhady2019}
M.~Abdelhady, A.~J. V.~D. Bogert, and D.~Simon, ``A high-fidelity wearable system for measuring lower-limb kinetics and kinematics,'' {\em IEEE Sensors Journal}, vol.~19, pp.~12482--12493, 12 2019.

\bibitem{Teufl2019}
W.~Teufl, M.~Miezal, B.~Taetz, M.~Frohlichi, and G.~Bleser, ``Validity of inertial sensor based 3d joint kinematics of static and dynamic sport and physiotherapy specific movements,'' {\em PLoS ONE}, vol.~14, 2 2019.

\bibitem{Sy2020}
L.~Sy, N.~H. Lovell, and S.~J. Redmond, {\em Estimating Lower Limb Kinematics Using a Lie Group Constrained Extended Kalman Filter with a Reduced Wearable IMU Count and Distance Measurements}.
\newblock MDPI Sensors, 2020.

\bibitem{Marimon2024}
X.~Marimon, I.~Mengual, C.~L. de~Celis, A.~Portela, J.~Rodríguez-Sanz, I.~A. Herráez, and A.~Pérez-Bellmunt, ``Kinematic analysis of human gait in healthy young adults using imu sensors: Exploring relevant machine learning features for clinical applications,'' {\em Bioengineering}, vol.~11, 2 2024.

\bibitem{Hossain2023}
M.~S.~B. Hossain, Z.~Guo, and H.~Choi, ``Estimation of lower extremity joint moments and 3d ground reaction forces using imu sensors in multiple walking conditions: A deep learning approach,'' {\em IEEE Journal of Biomedical and Health Informatics}, vol.~27, pp.~2829--2840, 6 2023.

\bibitem{Renani2021}
M.~S. Renani, A.~M. Eustace, C.~A. Myers, and C.~W. Clary, ``The use of synthetic imu signals in the training of deep learning models significantly improves the accuracy of joint kinematic predictions,'' {\em Sensors}, vol.~21, 9 2021.

\bibitem{Hernandez2021}
V.~Hernandez, D.~Dadkhah, V.~Babakeshizadeh, and D.~Kulić, ``Lower body kinematics estimation from wearable sensors for walking and running: A deep learning approach,'' {\em Gait and Posture}, vol.~83, pp.~185--193, 1 2021.

\bibitem{Chen2021}
Y.~L. Chen, I.~J. Yang, L.~C. Fu, J.~S. Lai, H.~W. Liang, and L.~Lu, ``Imu-based estimation of lower limb motion trajectory with graph convolution network,'' {\em IEEE Sensors Journal}, vol.~21, pp.~24549--24557, 11 2021.

\bibitem{Kaya2024}
E.~Kaya and H.~Argunsah, ``Exploring the contribution of joint angles and semg signals on joint torque prediction accuracy using lstm-based deep learning techniques,'' {\em Computer Methods in Biomechanics and Biomedical Engineering}, 2024.

\bibitem{Zhang2023}
J.~Zhang, Y.~Zhao, F.~Shone, Z.~Li, A.~F. Frangi, S.~Q. Xie, and Z.~Q. Zhang, ``Physics-informed deep learning for musculoskeletal modeling: Predicting muscle forces and joint kinematics from surface emg,'' {\em IEEE Transactions on Neural Systems and Rehabilitation Engineering}, vol.~31, pp.~484--493, 2023.

\bibitem{Song2023}
Q.~Song, X.~Ma, and Y.~Liu, ``Continuous online prediction of lower limb joints angles based on semg signals by deep learning approach,'' {\em Computers in Biology and Medicine}, vol.~163, 9 2023.

\bibitem{Shi2023}
Y.~Shi, S.~Ma, Y.~Zhao, and Z.~Zhang, ``A physics-informed low-shot learning for semg-based estimation of muscle force and joint kinematics,'' {\em archive}, 7 2023.

\bibitem{Truong2023}
M.~T.~N. Truong, A.~E.~A. Ali, D.~Owaki, and M.~Hayashibe, ``Emg-based estimation of lower limb joint angles and moments using long short-term memory network,'' {\em Sensors}, vol.~23, 3 2023.

\bibitem{Foroutannia2022}
A.~Foroutannia, M.~R. Akbarzadeh-T, and A.~Akbarzadeh, ``A deep learning strategy for emg-based joint position prediction in hip exoskeleton assistive robots,'' {\em Biomedical Signal Processing and Control}, vol.~75, 5 2022.

\bibitem{Wu2021}
W.~Wu, K.~R. Saul, and H.~Huang, ``Using reinforcement learning to estimate human joint moments from electromyography or joint kinematics: An alternative solution to musculoskeletal-based biomechanics,'' {\em Journal of Biomechanical Engineering}, vol.~143, 4 2021.

\bibitem{Zhang2021}
L.~Zhang, Z.~Li, Y.~Hu, C.~Smith, E.~M. Farewik, and R.~Wang, ``Ankle joint torque estimation using an emg-driven neuromusculoskeletal model and an artificial neural network model,'' {\em IEEE Transactions on Automation Science and Engineering}, vol.~18, pp.~564--573, 4 2021.

\bibitem{Wang2025}
Z.~Wang, C.~Chen, H.~Chen, Y.~Zhou, X.~Wang, X.~Wu, S.~Member, and X.~W.~Z. Wang, ``Dual transformer network for predicting joint angles and torques from multi-channel emg signals in the lower limbs,'' {\em IEEE Journal of Biomedical and Health Informatics}, 2025.

\bibitem{Vaswani2017}
A.~Vaswani, N.~Shazeer, N.~Parmar, J.~Uszkoreit, L.~Jones, A.~N. Gomez, Łukasz Kaiser, and I.~Polosukhin, ``Attention is all you need,'' {\em Advances in Neural Information Processing Systems}, vol.~2017-Decem, pp.~5999--6009, 2017.

\bibitem{Bahdanau2015}
D.~Bahdanau, K.~H. Cho, and Y.~Bengio, ``Neural machine translation by jointly learning to align and translate,'' {\em 3rd International Conference on Learning Representations, ICLR 2015 - Conference Track Proceedings}, pp.~1--15, 2015.

\bibitem{Luong2015}
M.~T. Luong, H.~Pham, and C.~D. Manning, ``Effective approaches to attention-based neural machine translation,'' {\em Conference Proceedings - EMNLP 2015: Conference on Empirical Methods in Natural Language Processing}, pp.~1412--1421, 2015.

\bibitem{Wang2023}
H.~Wang, A.~Basu, G.~Durandau, and M.~Sartori, ``A wearable real-time kinetic measurement sensor setup for human locomotion,'' {\em Wearable Technologies}, vol.~4, 2023.

\bibitem{He2015}
K.~He, X.~Zhang, S.~Ren, and J.~Sun, ``Deep residual learning for image recognition,'' {\em preprint}, 12 2015.

\bibitem{Abadi2016}
M.~Abadi, P.~Barham, J.~Chen, Z.~Chen, A.~Davis, J.~Dean, M.~Devin, S.~Ghemawat, G.~Irving, M.~Isard, M.~Kudlur, J.~Levenberg, R.~Monga, S.~Moore, D.~G. Murray, B.~Steiner, P.~Tucker, V.~Vasudevan, P.~Warden, M.~Wicke, Y.~Yu, and X.~Zheng, ``Tensorflow: A system for large-scale machine learning,'' {\em Proceedings of the 12th USENIX Symposium on Operating Systems Design and Implementation (OSDI ’16)}, p.~786, 11 2016.

\end{thebibliography}

\end{document}